\documentclass[conference]{ijdc-v14}
\usepackage{graphicx}
\usepackage{multirow}
\usepackage{makecell}
\usepackage{colortbl}
\usepackage[normalem]{ulem}
\useunder{\uline}{\ul}{}
\usepackage{amsmath}
\usepackage{xspace}
\usepackage{apalike}

\newcommand{\openrefine}{\textrm{OpenRefine}\xspace}

\usepackage{xcolor}

\definecolor{neonfuchsia}{rgb}{1.0, 0.25, 0.39}

\newcommand*{\affaddr}[1]{#1} 
\newcommand*{\affmark}[1][*]{\textsuperscript{#1}}
\newcommand*{\Scale}[2][4]{\scalebox{#1}{$#2$}}%

\title{T-KAER: Transparency-enhanced Knowledge-Augmented Entity Resolution Framework} 
\author{%
Lan Li\affmark[1], Liri Fang\affmark[1], Yiren Liu\affmark[2], Vetle I. Torvik\affmark[1], and Bertram Lud\"ascher\affmark[1,3]\\
\affaddr{\affmark[1]School of Information Sciences, University of Illinois Urbana Champaign}\\
\affaddr{\affmark[2]Informatics, University of Illinois Urbana Champaign}\\
\affaddr{\affmark[3]National Center for Supercomputing Applications (NCSA)}
}
\correspondence{Lan Li. Email: \email{\{lanl2\}@illinois.edu}}

\begin{document}

\maketitle

\begin{abstract}

Entity resolution (ER) is the process of determining whether two representations refer to the same real-world entity and plays a crucial role in data curation and data cleaning. Recent studies have introduced the KAER framework, aiming to improve pre-trained language models by augmenting external knowledge. However, identifying and documenting the external knowledge that is being augmented and understanding its contribution to the model's predictions have received little to no attention in the research community. This paper addresses this gap by introducing T-KAER, the \textbf{T}ransparency-enhanced \textbf{K}nowledge-\textbf{A}ugmented \textbf{E}ntity \textbf{R}esolution framework. 

To enhance transparency, three Transparency-related Questions (T-Qs) have been proposed: T-Q(1): What is the experimental process for matching results based on data inputs? T-Q(2): Which semantic information does \textit{KAER} augment in the raw data inputs? T-Q(3): Which semantic information of the augmented data inputs influences the predictions? To address the T-Qs, T-KAER is designed to improve transparency by documenting the entity resolution processes in log files. 

In experiments, a citation dataset is used to demonstrate the transparency components of T-KAER. This demonstration showcases how T-KAER facilitates error analysis from both quantitative and qualitative perspectives, providing evidence on ``what" semantic information is augmented and ``why" the augmented knowledge influences predictions differently.

\bigskip
\vspace{-0.4cm}
\end{abstract}

\vspace{-1cm}
\section{Introduction and Overview}

The FAIR guiding principles for scientific data aim to ensure that data is \emph{findable}, \emph{accessible}, \emph{interoperable}, and \emph{reusable} \cite{wilkinson_fair_2016}. The concept of research \emph{transparency} follows as one way to improve reusability of data and the reproducibility of results \cite{nosek2015promoting,mcphillips2019reproducibility}. In addition to reproducibility, transparency in the research process is always essential to check research integrity, identify fraud, and track retractions~\cite{lyon2016transparency}. Essentially, a transparent research process leads to greater trustworthiness by enabling researchers to easily track and verify internal products and understand mechanisms, even without re-running.

In this paper, we explore transparency in entity resolution (ER), the problem of determining whether two separate representations refer to the same real-world entity, regardless of whether they exist within the same database or span across different databases \cite{christen_data_2012}.
ER helps reconcile data inconsistencies and eliminate duplicates during data integration, playing a crucial role in data curation. Consequently, there is an increasing demand for a reliable and user-friendly entity resolution tool that minimizes the effort required by data curators. ER is also referred to as \emph{deduplication} \cite{koumarelas_data_2020}. Existing data cleaning tools, such as \openrefine~\cite{openrefine2020}, use traditional machine learning algorithms like K-Nearest Neighbors (KNN) to detect duplications. Nowadays, deep learning techniques \cite{arabnia_when_2021}, pre-trained language models (PLMs)
\cite{li_deep_2020,li_improving_2021,paganelli_analyzing_2022}, and even large language models (LLMs) \cite{peeters2023using}, have been deployed to tackle entity resolution. 

\textbf{Why augment PLMs with domain knowledge for ER?} Many existing methods of entity resolution hypothesize that records of data follow the same known schema \cite{elmagarmid_duplicate_2007}. However, this is not always the case in real-world applications. Raw data is often collected from sources that are highly heterogeneous and do not share a common schema. The source data can also come from multiple domains, and it may be represented in diverse formats. The complexity involved makes it challenging for data curators to perform entity resolution without specialized knowledge about the data's domain. For instance, a pair of records from the citation domain might include title, author names, venue name, and publication year. The potential challenges for the model to understand citation data include: (1). Not all attributes hold equal importance; title, venue name, and publication year matter more than author names. (2). There are two layers of semantic information in author names: values and order. Even if the spellings for author names are the same, different orders might result in a \textbf{not-match}. Additionally, authors' names may be presented in different formats based on the citation type, such as the position of the first name and last name or acronyms of the names. Hence, it is beneficial for entity resolution methods to integrate additional semantic information into the source data.

\textbf{What are the challenges of using PLMs in ER?} As emphasized in \cite{li2021deep}, the responsible management of data requires that algorithms used in entity resolution tasks be explainable: This means, it is particularly critical for comprehending the reasons behind matching entities and, equally important, why certain entities are not considered a match. One shared challenge faced by applications based on these PLMs is their ``black-box'' nature. Despite certain transformer-based language models having open-access architectures, the lack of transparency in their pretraining data and processes can lead to a loss of understanding regarding why various data inputs resulted in the final matching results. 

\textbf{How to enable Transparency in ER?} As mentioned in \cite{grafberger2023provenance}, ``provenance is all you need'', suggesting that enabling provenance tracking can automate the detection of many common correctness issues in machine learning pipelines. Specifically, they emphasize the use of \emph{why-provenance} to determine which data inputs were used to compute specific data outputs. Furthermore, we invite the concept of \emph{where-provenance} \cite{buneman2001and} to describe which semantic information of the augmented data inputs influences the entity resolution results. By explaining three T-Qs related to the augmentation of additional domain knowledge for matching results, we aim to provide a more transparent entity resolution process.

\begin{figure*}[!ht]
    \centering
    \includegraphics[width=\linewidth]{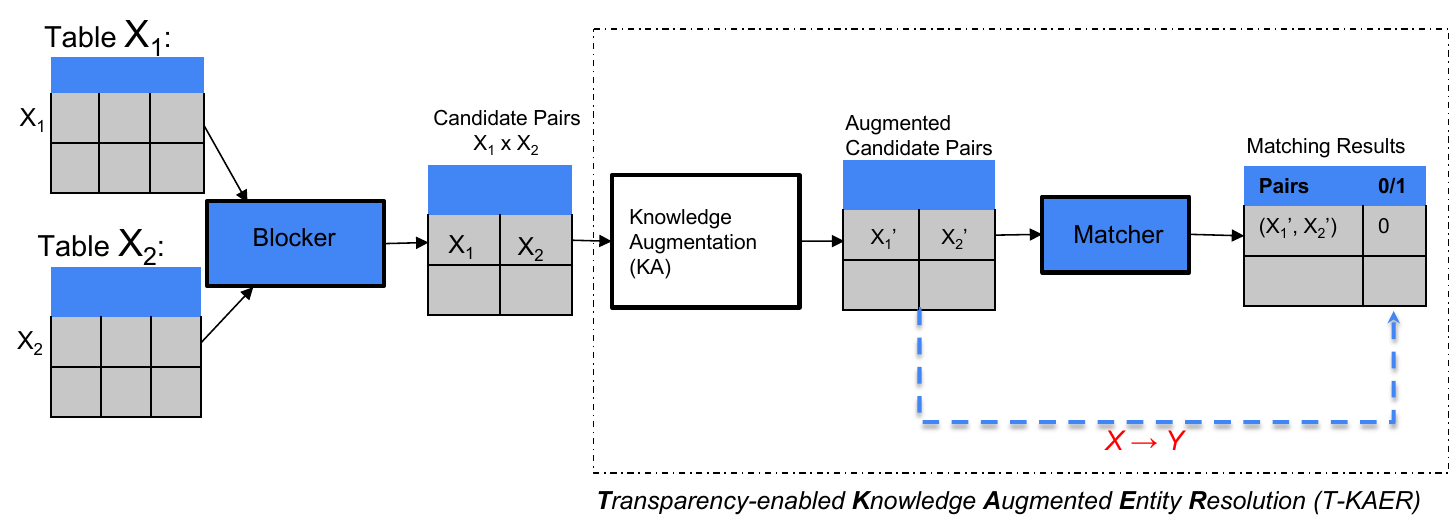}
    \caption{Table $X_1$ and Table $X_2$ are a pair of relational tables. Candidate pairs $(x_1, x_2) \in X_1 \times X_2$ are executed by the \textbf{Blocker}. Then follows the pipeline of Knowledge Augmented Entity Resolution (\textit{KAER} framework): Augmented candidate pair $(x_1', x_2')$ is processed by KA component. Then it is passed to the \textbf{Matcher} and returns the matching results (0: not-match,1:match) to data curators. Transparency is enhanced through providing evidence explaining how the augmented data inputs influence the decision-making process.}
    \label{fig:tkaer}
\end{figure*}

\textbf{Contributions} We introduce \textit{T-KAER} \footnote{T-KAER is public and freely available from GitHub: https://github.com/idaks/knowledge-augmented-entity-resolution} (Figure \ref{fig:tkaer}) to enhance the transparency for entity resolution process. In summary, this paper makes three main contributions:

\begin{itemize}
    \item Proposing and addressing three transparency-related questions (T-Qs) to enhance transparency in the Knowledge-Augmented Entity Resolution (KAER \cite{fang2023kaer}) framework .
    \item Designing a provenance-persevering pipeline enables modeling the training information into the structured log files, with the aim of supporting error analysis from both quantitative and qualitative perspectives.
    \item Conducting experiments on a citation dataset to demonstrate how T-KAER can facilitate error analysis and enhance transparency.
\end{itemize}

\vspace{-0.3cm}
\section{Notation Definitions and Related Work}
\vspace{-0.3cm}
\subsection{Notation of Entity Resolution} 

The input of the entity resolution task consists of a set $M \subseteq X_1 \times X_2$, where $X_1$ and $X_2$ are two sets of data entry collections that contain duplicated entries. Each data entry, $x_1 \in X_1$ or $x_2 \in X_2$, is formatted in ${(col_i, val_i)}_{1 \leq i \leq N}$, containing $N$ column and values. The task discussed in this paper focuses on: for each data entry pair $(x_1, x_2) \in M$, determine whether $x_1$ and $x_2$ refer to the same data entity.
\vspace{-0.5cm}
\subsection{KAER: Pre-trained Language Model for Entity Resolution}
A few recent works apply transformer-based PLMs to entity resolution tasks. \cite{paganelli_analyzing_2022} discover that simply fine-tuning BERT can benefit matching/not-matching classification tasks and recognize the input sequence as a pair of records. Ditto by \cite{li_deep_2020} is the state-of-the-art entity matching system based on PLMs, i.e., RoBERTa. In addition, Ditto provides a deeper language understanding for entity resolution by injecting domain knowledge, summarizing the key information, and augmenting with more difficult examples for training data. Following the work by~\cite{li_deep_2020}, KAER uses RoBERTa as the backbone model.

In summary, KAER uses PLMs for entity resolution and contains three modules for knowledge augmentation: a) column semantic type augmentation, b) entity semantic type augmentation, and c) three options of prompting types. The subsequent sections will describe each module within KAER.


\textbf{Column-Level Knowledge Augmentation}. Semantic column-type augmentation can inject domain-specific knowledge for columns with/without existing column names. Existing studies \cite{hulsebos_sherlock_2019,suhara_annotating_2022} use deep learning approaches to detect semantic data types at the column level. We adapt existing methods: \textit{Sherlock} \cite{hulsebos_sherlock_2019} and \textit{Doduo} \cite{suhara_annotating_2022}, to perform column semantic typing prediction. \textit{Sherlock} is a multi-input deep neural network that uses multiple feature sets, including embeddings and column statistics, with a multi-layer sub-neural network applied to each column-wise feature set, and the output is fed into a primary neural network. Comparably, \textit{Doduo} is a multi-task learning framework based on PLMs \cite{suhara_annotating_2022}. \textit{Doduo} serializes the entire table into a sequence of tokens, i.e., concatenating column values to make a sequence of tokens and feed that sequence as input to the transformer \cite{suhara_annotating_2022}. 

\textbf{Entity-Level Knowledge Augmentation}. Entity semantic type augmentation leverages the entity linking \cite{li_deep_2020} method to identify all entity mentions from a given knowledge base (KB) within a given text input.  Entity linking (EL) refers to linking entity mentions appearing in natural language text with their corresponding entities in an external knowledge base, e.g., Wikidata. Ayoola et al.~\cite{ayoola_refined_2022} introduced an EL method by fine-tuning a PLM (RoBERTa) over Wikidata, which is used for EL in this study, which will be examined in this work as the entity-level knowledge augmentation methods. 

\textbf{Prompting Types}. KAER has employed two different methods of prompting for knowledge augmentation, template-based and constrained tuning. For template-based prompting, text-based templates are utilized to verbalize the domain knowledge as text input, using connectors such as slash\footnote{We use the slash to represent the ``or'' semantic that is commonly present in the general web corpus.} (``/'') and space. Second, we have examined prompting with constrained tuning by employing soft-position encoding and visible matrix, informed by the method introduced in K\-Bert~\cite{liu_k-bert_2020}.

\section{T-KAER Introduction}

\begin{figure*}[!th]
    \centering
    \includegraphics[width=0.6\linewidth]{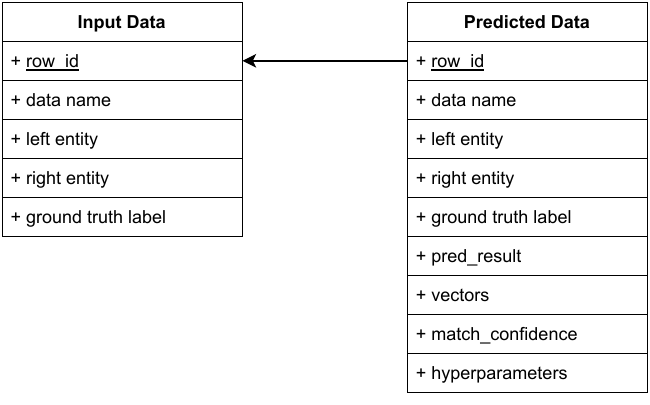}
    \caption{The UML diagram showing the structure of log file in JSON format. 
    The \textbf{Input Data} collects information before running the experiment, and the \textbf{Predicted Data} appends run-time parameters and computation results to the \textbf{Input Data} after the experiment.
    }
    \label{fig:tkaer-struct}
    \vspace{-0.4cm}
\end{figure*}

T-KAER enhances the transparency of the entity resolution process by documenting the experimental process into a log file. A list of variables applied and data products generated during the experimental process are recorded (see table \textbf{Predicted Data} from Figure \ref{fig:tkaer-struct}). In the subsequent sections, we will introduce the main components for T-KAER, and using the datalog queries, a declarative programming language, to show how the system models each transparency-related question (T-Q) with the recorded information.

\subsection{Entry Inputs}
For each entity pair, $(e_1, e_2)$, the text context of column names and values of $e_1$ and $e_2$ are serialized and concatenated as the input for PLMs. The [CLS] token position is used to classify whether $e_1$ and $e_2$ refer to the same entity. The loss for optimizing the classification objective is:

\begin{equation}
    \ell = - log\ p(y | s(e_1, e_2))
\end{equation}

where y denotes whether $e_1$ and $e_2$ refer to the same entity, and $s(\cdot,\cdot)$ denotes the serialization and transformation of entity pairs with knowledge injection and prompting methods. 
\begin{equation}
\Scale[0.8]{
    s(e_i, e_j) ::=  \text{[CLS]}\ serialize(e_i)\ \text{[SEP]}\ serialize(e_j)\ \text{[SEP]}} 
\end{equation}
where $serialize(\cdot)$ serializes each data entry.
\begin{equation}\Scale[0.8]{
\begin{aligned}
serialize(e_i) ::= & \text{[COL]}\ f(col_1,pt)\ \text{[VAL]}\ g(val_1,pt)\ ...\\
            & \text{[COL]}\ f(col_N,pt)\ \text{[VAL]}\ g(val_N,pt)   
\end{aligned}}
\label{eq:serialize}
\end{equation}
where $f(col_i, pt)$ denotes the semantic column type injection with prompting method $pt$, and $g(val_i, pt)$ denotes the EL injection with prompting method $pt$. 

\vspace{-0.5cm}
\subsection{Contents of Log Files}

Figure \ref{fig:tkaer-struct} showcases the information collected in a log file. Before the experiment, the dataset name, and for each row, the row index, entry inputs (left entity and right entity), and the ground truth are recorded in the \textbf{Input Data}. After running the experiment, the system will harvest the variables used and internal products computed during the testing process into the \textbf{Predict Data}. This includes hyperparameters, embedding vectors for each row of entry inputs by the model, predicted results (\textbf{match} or \textbf{not-match}), and matching confidence.

\vspace{-0.5cm}
\subsection{Three Transparency-related Questions (T-Qs) Answered by Datalog Queries}

\begin{figure*}[!ht]
    \centering
    \includegraphics[width=0.5\linewidth]{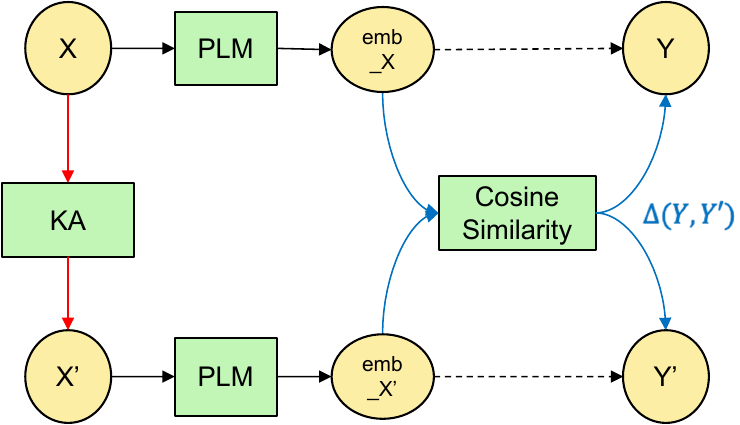}
    \caption{Yellow nodes represent the data and variables, green boxes represent the processes, such as KA for the knowledge augmentation process. Modeling three T-Qs with log files: T-Q(1) - Horizontally, the black path showcases various entry inputs X and X' resulting in predicted results Y and Y'; T-Q(2) - The vertical red path (X → X') tracks how KA methods alter the entry inputs; T-Q(3) - The blue curve path represents the cosine similarity between embeddings, determining the similarity of predicted results. 
    }
    \label{fig:transparency_model}
\end{figure*}

In this section, we will model and explain three transparency-related questions (T-Qs) (See Figure \ref{fig:transparency_model}). We will use Datalog~\cite{ceri1989you} queries to represent the retrieval process. The tables \textbf{Input Data} and \textbf{Predicted Data} will be used to address the T-Qs.

\vspace{-0.3cm}

\vspace{-0.5cm}

\subsubsection{T-Q(1). What is the experimental process for matching results based on data inputs?}

This question requires documenting the experimental process by retrieving entity pairs as entry inputs and predicted results by each method.

\begin{equation}\Scale[0.8]{
X\_to\_Y(A, ``Sherlock", L, R, G, Y) :- Predicted\_Data(A, ``Sherlock", L, R, G, Y, \_, \_, \_).                         }
\end{equation}

Denote that \textbf{Predicted\_Data} (recorded in Figure \ref{fig:tkaer-struct}) is needed to retrieve left (L) and right entity (R), ground truth label (G), and predicted result (Y) every row and by various methods. Here, the method name is ``Sherlock". 
\vspace{-0.3cm}

\subsubsection{T-Q(2). Which semantic information does KAER augment in the raw data inputs?}

By exploring this question, we can compare various entry inputs augmented by different methods. As mentioned above, there are two types of knowledge: column level, and entity level (See the examples in Table \ref{Tab:knowledge_augmentation_examples}). Various knowledge augmented into the entity pairs (left entity and right entity) will result in the difference between the semantic information contained. 

To address T-Q(2), \textbf{Input Data} (recorded in Figure \ref{fig:tkaer-struct}) is used: 
\vspace{-0.2cm}
\begin{equation}\Scale[0.8]{
\begin{aligned}
delta\_X(A1, ``Sherlock", L1, R1, ``Doduo", L2, R2) :- & Input\_Data(A1, ``Sherlock", L1, R1, \_), \\ 
                                                          & Input\_Data(A1, ``Doduo", L2, R2, \_).
\end{aligned}}
\vspace{-0.2cm}
\end{equation}

Denote that L1 and R1 are the left entity and right entity from ``original" dataset, while L2 and R2 are from ``Sherlock"-augmented dataset. This equation is to help compare entity pairs augmented by different methods.

\begin{table}[!ht]
\small
\caption{Semantic information added: column semantic typing, and entity linking. In column \textit{column name/ column semantic typing}: the first row is predicted by Sherlock \cite{hulsebos_sherlock_2019}, in which the \textit{``name''} is augmented with semantic type \textit{``song\_name''}, and the second row is predicted by Doduo \cite{suhara_annotating_2022}, in which \textit{``title''} can be augmented with \textit{``computer.software''}. In column \textit{cell value / entity linking}, entity mentions such as \textit{``Illusion"} is annotated as \textit{``single"}, and \textit{``protocol"} is identified as \textit{``computer network protocol"}.}
\label{Tab:knowledge_augmentation_examples}
\resizebox{\textwidth}{!}{%
\begin{tabular}{|p{1.5cm}|p{8cm}|p{2.5cm}|p{8cm}|}
\hline
\textbf{column name (original)} & \textbf{cell value (original)} & \textbf{column name / column semantic typing}& \textbf{cell value /  entity linking}\\
\hline
name & Illusion ( feat . Echosmith ) Zedd True Colors Dance, Music, Electronic 2015 Interscope Records 6:30 & name \textcolor{red}{/ song\_name} & Illusion \textcolor{red}{/ single} ( feat . Echosmith ) Zedd True Colors Dance, Music, Electronic 2015 Interscope Records 6:30 \\
\hline
title & the demarcation protocol: a technique for maintaining constraints in distributed database systems vldb j. 1994 & title \textcolor{red}{/ computer.software} & the demarcation protocol \textcolor{red}{/ computer network protocol}: a technique for maintaining constraints in distributed database systems vldb j. 1994 \\
\hline
\end{tabular}
}
\end{table}
\vspace{-0.3cm}
\subsubsection{T-Q(3). Which semantic information of the augmented inputs influences the predictions?}

This question aims to explore the impact of augmenting entity pairs using various knowledge augmentation methods on the differences in predicted results. Specifically, by incorporating domain knowledge at the column level, entity level, or both levels, the predicted results may either improve or worsen.

Compared to T-Q(2), which retrieves various entry inputs in text, T-Q(3) compares the embedding vectors based on entry inputs generated by PLMs. In detail, the system collects embedding vectors that yield the same correct predicted result and embedding vectors that yield different predicted results. Then the similarity between embedding vectors under each condition is compared to determine whether the former is higher than the latter. The assumption is that embedding vectors can reflect semantic information


Situation I: Various augmentation methods result in the different predicted results:
\begin{equation}\Scale[0.8]{
\begin{aligned}
delta\_Y(A1, ``Sherlock", L1, R1, Y1, V1, ``Doduo", L2, R2, Y2, V2, G) :- \\
        Predicted\_Data(A1, ``Sherlock", L1, R1, G, Y1, V1, \_, \_), \\ 
        Predicted\_Data(A1, ``Doduo", L2, R2, G, Y2, V2, \_, \_), \\
        Y1 != Y2.
\end{aligned}}
\vspace{-0.2cm}
\end{equation}

Situation II: Various augmentation methods result in the same and correct predicted results:
\begin{equation}\Scale[0.8]{
\begin{aligned}
delta\_Y(A1, ``Sherlock", L1, R1, Y1, V1, ``Doduo", L2, R2, Y2, V2) :- \\
        Predicted\_Data(A1, ``Sherlock", L, R, G, Y, \_, \_, \_), \\ 
        Predicted\_Data(A1, ``Sherlock", L, R, G, Y, \_, \_, \_), \\
        Y1=Y2, Y1=G.
\end{aligned}}
\vspace{-0.6cm}
\end{equation}

\section{Experiment and Results Analysis}

KAER is evaluated on the Magellan datasets \cite{magellandata} across various domains. In this section, we will introduce one of the datasets as an example for a case study, namely the \textit{DBLP-ACM} dataset from the citation domain. Then, we will describe the experimental settings and results analysis in both quantitative and qualitative analysis.  

\subsection{Dataset Description}
\textit{DBLP-ACM} dataset comprises four attributes: title, authors, venue, and year. Each entry represents a publication record, with the title indicating the paper or article name, authors containing the author names, and the venue specifying the platform or journal where the publication is released. The data input of each entity pair is the serialized string following Eq.~\ref{eq:serialize}. That is, for each entity pair column names and values will be concatenated and serialized into one single string. For instance, `Entity 1' at row id 1457 from Table \ref{tab:original-ds} will be serialized as: ``COL \textit{title} VAL \textit{a formal perspective on the view selection problem} COL \textit{authors} VAL \textit{rada chirkova, dan suciu, alon y. halevy } COL \textit{venue} VAL \textit{vldb j.} COL \textit{year} VAL \textit{2002}" , and `Entity 2' will be serialized as ``COL \textit{title} VAL \textit{a formal perspective on the view selection problem} COL authors VAL \textit{rada chirkova, alon y. halevy, dan suciu} COL venue VAL \textit{very large data bases} COL \textit{year} VAL \textit{2002}". Entities 1 and 2 will be concatenated by space and used as data input for PLMs. 




\subsection{Experimental Settings}

As illustrated previously, there are two levels of knowledge augmentation (KA) methods: column level and entity level. Consequently, we established two groups of datasets for experiments: one with column-level KA datasets only and another with combined (combination of column-level and entity-level) KA datasets (See Table \ref{tab:tkaer-exp}). 
\begin{table}[!ht]
    \centering
    \caption{Experiments are categorized into two groups: Column-level KA datasets, and Column-Entity level KA datasets.}
    \vspace{-0.3cm}
    \resizebox{\textwidth}{!}{
    \begin{tabular}{|p{1.5cm}|p{10cm}|p{10cm}|} \hline 
         Types&  Experiment I (Column Level)& Experiment II (Column Level \& Entity Level) \\ \hline 
         Test I&  Matching results by Sherlock and Doduo are different& Matching results by Doduo and Doduo with EL (Entity Linking by ReFinED) are different\\ \hline 
         Test II&  Matching results by Sherlock and Doduo are the same and correct (predicted results by non-KA are wrong)& Matching results by Doduo and Doduo with EL are the same and correct (On the condition that the predicted results by non-KA are incorrect)\\ \hline
    \end{tabular}
    }
    \label{tab:tkaer-exp}
    \vspace{-0.5cm}
\end{table}

\subsection{Results Analysis}
Documenting the experimental process into log files facilitates the retrieval of information to address three T-Qs. In a nutshell, T-Q(1) compares various predicted results, T-Q(2) compares various entry inputs, and T-Q(3) examines the similarity of various embedding vectors. The results analysis will be presented based on the log files from two perspectives: quantitative and qualitative.

\subsubsection{Quantitative Analysis}

There are two experiments processed for quantitative analysis: (1) Count how many rows fulfill the requirements in Test I and Test II. (2) Compare cosine similarity between embedding vectors based on various entry inputs in Test I and Test II (See conditions for Test I and Test II in Table \ref{tab:tkaer-exp}). 
\paragraph{Evaluate Performance of KA Methods for Predicted Results: T-Q(1)}

This evaluation result (see Table \ref{tab:tkaer-qa1}) helps address T-Q(1), the number of rows that the predicted matching results are influenced by KA methods. The left table illustrates the performance of column-level augmented methods only, while the right table compares column-level (i.p. KA by Doduo) and combined augmented methods. 

In Experiment I, where column semantic typing is injected by both methods, three rows of matching results show improvement. Sherlock correctly predicts three rows that Doduo misclassifies, and Doduo correctly predicts two rows that Sherlock misclassifies. 

For Experiment II, both methods enhance the performance of matching result prediction by two rows. The combined method improves five rows of predicted results misclassified by Doduo. Paradoxically, there are four rows correctly predicted by Doduo but misclassified by the combined method.

\begin{table}[!ht]
\centering
\vspace{-0.4cm}
\caption{Result analysis on T-Q(1). Left Table: compares prediction results of column-level augmented methods (Sherlock and Doduo). Right Table: compares prediction results of column-level (Doduo) and combined augmented method (Doduo \& EL). T: Predicted results equal to the ground truth; F: Predicted results do not equal the ground truth. We compute the number of rows and the ratio (=row count/ total number of rows). The last row for each table represents cases where both KA methods yield true results, given a false prediction by non-KA.}
\label{tab:tkaer-qa1}
\resizebox{\textwidth}{!}{%
\begin{tabular}{|p{3cm}|p{3cm}|p{2cm}|p{1.5cm}|p{0.1cm}|p{3cm}|p{3cm}|p{2cm}|p{1.5cm}|}
\cline{1-4} \cline{6-9}
\multicolumn{2}{|l|}{Predicted Results} &
  \multirow{2}{*}{Row Count} &
  \multirow{2}{*}{Ratio} &
   &
  \multicolumn{2}{c|}{Predicted Results} &
  \multirow{2}{*}{Row Count} &
  \multirow{2}{*}{Ratio} \\ \cline{1-2} \cline{6-7}
\multicolumn{1}{|l|}{Sherlock} & Doduo &   &        &  & \multicolumn{1}{l|}{Doduo} & Doduo \& EL &   &        \\ \cline{1-4} \cline{6-9} 
\multicolumn{1}{|l|}{T}        & F     & 3 & 0.0012 &  & \multicolumn{1}{l|}{T}     & F           & 4 & 0.0016 \\ \cline{1-4} \cline{6-9} 
\multicolumn{1}{|l|}{F}        & T     & 2 & 0.0008 &  & \multicolumn{1}{l|}{F}     & T           & 5 & 0.0020 \\ \cline{1-4} \cline{6-9} 
\multicolumn{1}{|l|}{T}        & T     & 3 & 0.0012 &  & \multicolumn{1}{l|}{T}     & T           & 2 & 0.0008 \\ \cline{1-4} \cline{6-9} 
\end{tabular}%
}
\vspace{-0.5cm}
\end{table}

\paragraph{Evaluate Semantic Information Contained in Embedding Vectors: T-Q(3)}

The evaluation results (see Table \ref{tab:tkaer-qa2}) address T-Q(3), illustrating how internal products, embedding vectors generated by PLMs to reflect the semantic information in entry inputs, lead to either identical or different predicted results. We calculate the cosine similarity for embeddings in Test I and Test II, comparing the average similarity for each test.

The left table \ref{tab:tkaer-qa2} showcases the column-level methods, while the right table compares column-level and combined methods. In both experiments, the average cosine similarity in Test II is higher than in Test I. Embedding vectors in Test II are derived from entry inputs through various augmentation methods, resulting in the same prediction result. On the other hand, embedding vectors in Test I are generated from entry inputs leading to different prediction results. The higher cosine similarity between embeddings represents they contain more similar semantic information. This aligns with our assumption that ``Data inputs resulting in the same predicted results (Test II) exhibit more similar embeddings (with similar semantic information) compared to those resulting in different predicted results (Test I)". This holds true on average.

\begin{table*}[ht!]
\centering
\caption{Result analysis on T-Q(3). Left Table: compares prediction results of column-level augmented methods (Sherlock and Doduo). Right Table: compares predictions of column-level (Doduo) combined augmented methods (Doduo \& EL). T: Predicted results equal to the ground truth; F: Predicted results do not equal to the ground truth. We calculate the average of cosine similarity (Avg.) for embeddings for Test I and Test II.}
\label{tab:tkaer-qa2}
\resizebox{\textwidth}{!}{%
\begin{tabular}{|p{1.5cm}|p{3cm}|p{3cm}|m{2.5cm}|p{2cm}|p{0.2cm}|p{1.5cm}|p{3cm}|p{3cm}|m{2.5cm}|p{2cm}|}
\cline{1-5} \cline{7-11}
\multirow{2}{*}{Test ID} & \multicolumn{2}{c|}{Predicted Results} & \multirow{2}{*}{Ground Truth} & \multirow{2}{*}{Avg} &  & \multirow{2}{*}{Test ID} & \multicolumn{2}{c|}{Predicted Results}   & \multirow{2}{*}{Ground Truth} & \multirow{2}{*}{Avg} \\ \cline{2-3} \cline{8-9}
                         & \multicolumn{1}{c|}{Sherlock}  & Doduo &                               &                      &  &                          & \multicolumn{1}{c|}{Doduo} & Doduo \& EL &                               &                      \\ \cline{1-5} \cline{7-11} 
Test I                   & \multicolumn{1}{c|}{T}         & F     & 1                             & 0.4980               &  & Test I                   & \multicolumn{1}{c|}{T}     & F           & 1                             & 0.2749               \\
                         & \multicolumn{1}{c|}{F}         & T     & 1                             &                      &  &                          & \multicolumn{1}{c|}{F}     & T           & 1                             &                      \\
                         & \multicolumn{1}{c|}{T}         & F     & 0                             &                      &  &                          & \multicolumn{1}{c|}{T}     & F           & 0                             &                      \\
                         & \multicolumn{1}{c|}{F}         & T     & 0                             &                      &  &                          & \multicolumn{1}{c|}{F}     & T           & 0                             &                      \\ \cline{1-5} \cline{7-11} 
Test II                  & \multicolumn{1}{c|}{T}         & T     & 1                             & 0.6475               &  & Test II                  & \multicolumn{1}{c|}{T}     & T           & 1                             & 0.5796               \\
                         & \multicolumn{1}{c|}{T}         & T     & 0                             &                      &  &                          & \multicolumn{1}{c|}{T}     & T           & 0                             &                      \\ \cline{1-5} \cline{7-11} 
\end{tabular}
}
\vspace{-0.4cm}
\end{table*}



        

\subsubsection{Qualitative Analysis}

To gain a deeper understanding of how KA methods influence entity resolution results, we select certain entity pairs from the quantitative analysis and proceed with qualitative analysis. The chosen entity pairs from the original dataset are listed in Table \ref{tab:original-ds}. Next, we manually complete the entity resolution tasks for the selected rows, highlighting potential challenges. Subsequently, we present the predicted results on those rows using various KA methods and provide explanatory reasons through error analysis based on T-KAER. Therefore, the log files enable us to explain T-Q(1) and T-Q(2) explicitly in these case studies.  

\begin{table}[!ht]
    \centering
     \caption{Selected rows from the original \textbf{DBLP-ACM} dataset for error analysis. `Entity 1' is from source DBLP. `Entity 2' is from source ACM, highlighted in grey.}
     \vspace{-0.4cm}
    \label{tab:original-ds}
    \resizebox{\textwidth}{!}{
    \begin{tabular}{|p{1.4cm}|p{1.2cm}|p{7cm}|p{8cm}|p{4cm}|p{1cm}|} \hline
    \textbf{entry} & \textbf{row id} & \textbf{title}                                      & \textbf{authors}  & \textbf{venue}                           & \textbf{year} \\ \hline
    Entity 1       & 2437            & the mariposa distributed database management system & jeff sidell       & sigmod record                       & 1996          \\ \hline
    \rowcolor[HTML]{EFEFEF} 
    Entity 2 &
      2437 &
      mariposa : a wide-area distributed database system &
      michael stonebraker , paul m. aoki , witold litwin , avi pfeffer , adam sah , jeff sidell , carl staelin , andrew yu &
      the vldb journal -- the international journal on very large data bases &
      1996 \\ \hline
    Entity 1 &
      2407 &
      a formal perspective on the view selection problem &
      rada chirkova , dan suciu , alon y. halevy &
      vldb j. &
      2002 \\ \hline
    \rowcolor[HTML]{EFEFEF} 
    Entity 2 &
      2407 &
      a formal perspective on the view selection problem &
      rada chirkova , alon y. halevy , dan suciu &
      very large data bases &
      2001 \\ \hline
    Entity 1 &
      1457 &
      reminiscences on influential papers &
      hector garcia-molina , patricia g. selinger , tomasz imielinski , david maier , jeffrey d. ullman , richard t. snodgrass &
      sigmod record &
      1998 \\ \hline
    \rowcolor[HTML]{EFEFEF} 
    Entity 2       & 1457            & reminiscences on influential papers                 & richard snodgrass & acm sigmod record & 1998          \\ \hline
    \end{tabular}%
}
\vspace{-0.4cm}
\end{table}

\paragraph{Entity Resolution Finished Manually}

Entity pairs at row 2437 (the first two rows in Table~\ref{tab:original-ds}) are evidently not a match, as the values for title and venue differ significantly. Additionally, the authors are not identical. In the case of entity pairs at row 2407, determining a match might be challenging without domain knowledge in citation data, as the order of author names and the publication year are crucial. Therefore, despite these two entities sharing the same title, the same set of author names, and a similar meaning of venue name (i.e., `vldb j.' is the acronym for `very large data bases'), entity pairs at row 2407 are marked as \textbf{not-match}. For entity pairs at row 1457, we observe that the values for key columns ``title," ``venue," and ``year" are the same, but the author names are not exactly identical. It turns out that this record contains scientific commentaries on various works by different authors. Therefore, the single editor name: \textit{Richard Snodgrass} is sufficient to represent other authors. Consequently, the last pair of entities refers to the same citation.

\paragraph{Error Analysis Supported by T-KAER}

\textbf{Case Study I: Entity Resolution Results by Column-Level KA Methods are True.}

Entity pairs at row 2437 are correctly predicted as \textbf{not-match} by both Sherlock and Doduo. The entry inputs augmented by these two methods are as follows (see Table \ref{tab:ea-1}). Note that the column semantic types (CST) predicted by the same method for Entity 1 and Entity 2 are even slightly different, which is mainly due to the separate training of two datasets (i.e., DBLP and ACM datasets). Overall, the column semantic types predicted by Doduo are more precise than those predicted by Sherlock. For instance, values in the column ``authors" are recognized as ``people.person," and the year is annotated as ``time.event." Despite the poor CST predicted by Sherlock, the entity resolution results from both methods are correct. One reason is that most of the strings in the sequence are different, so it does not significantly impact the predicted results, regardless of whether the augmented knowledge is correct or not.

\begin{table}[!ht]
    \centering
    \caption{Augmented Entity pairs at row 2437: The CST augmented by these two methods is highlighted in \textbf{bold}.}
    \vspace{-0.4cm}
    \label{tab:ea-1}
    \resizebox{\textwidth}{!}{
    \begin{tabular}{|p{2cm}|p{8cm}|p{8cm}|} \hline 
         Entity Pair &  entry\_sherlock& entry\_doduo\\ \hline 
         Entity 1&  COL title \textbf{symbol} VAL the mariposa distributed database management system COL authors \textbf{area} VAL jeff sidell COL venue \textbf{person} VAL sigmod record COL year \textbf{education} VAL 1996 & COL title \textbf{business.industry} VAL the mariposa distributed database management system COL authors \textbf{people.person} VAL jeff sidell COL venue \textbf{organization.organization} VAL sigmod record COL year \textbf{time.event} VAL 1996 \\ \hline 
 Entity 2& COL title \textbf{result} VAL mariposa : a wide-area distributed database system COL authors \textbf{category} VAL michael stonebraker , paul m. aoki , witold litwin , avi pfeffer , adam sah , jeff sidell , carl staelin , andrew yu COL venue \textbf{code} VAL the vldb journal -- the international journal on very large data bases COL year \textbf{education} VAL 1996
&COL title \textbf{business.industry} VAL the mariposa distributed database management system COL authors \textbf{people.person }VAL jeff sidell COL venue \textbf{organization.organization} VAL sigmod record COL year \textbf{time.event} VAL 1996 \\ \hline
    \end{tabular}
    }
    
\end{table}

\textbf{Case Study II: Entity Resolution Results Predicted by Column-Level KA Methods are Different.}

These two records are incorrectly predicted as \textbf{match} by Sherlock, but correctly predicted as \textbf{not-match} by Doduo. While comparing the entry inputs, there are significant differences between the column semantic types augmented by Sherlock and Doduo. Column ``authors" is predicted as ``area" by Sherlock, correctly recognized as ``people.person" by Doduo. For column ``venue", Sherlock predicts the column values as ``person", on the contrary, Doduo labels it as ``organization.organization" and ``book.periodical". Finally, the column ``year" is annotated as ``education" by Sherlock, as ``time.event" by Doduo instead. Consequently, knowledge augmented by Doduo is more precise than Sherlock.

\begin{table}[!ht]
    \centering
    \caption{Augmented Entity pairs at row 2407: The CST augmented by these two methods is highlighted in \textbf{bold}.}
    \vspace{-0.4cm}
    \label{tab:ea-2}
    \resizebox{\textwidth}{!}{
    \begin{tabular}{|p{2cm}|p{8cm}|p{8cm}|} \hline 
         Entity Pair &  entry\_sherlock& entry\_doduo\\ \hline 
         Entity 1&  COL title \textbf{symbol} VAL a formal perspective on the view selection problem COL authors \textbf{area} VAL rada chirkova , dan suciu , alon y. halevy COL venue \textbf{person} VAL vldb j. COL year \textbf{education} VAL 2002 & COL title \textbf{business.industry} VAL a formal perspective on the view selection problem COL authors \textbf{people.person} VAL  rada chirkova , dan suciu , alon y. halevy COL venue \textbf{organization.organization} VAL vldb j. COL year \textbf{time.event} VAL 2002 \\ \hline 
 Entity 2& COL title \textbf{result} VAL a formal perspective on the view selection problem COL authors \textbf{category} VAL rada chirkova , alon y. halevy , dan suciu COL venue \textbf{code} VAL very large data bases COL year \textbf{education} VAL 2001
&COL title \textbf{business.industry} VAL a formal perspective on the view selection problem COL authors \textbf{people.person }VAL rada chirkova , alon y. halevy , dan suciu  COL venue \textbf{book.periodical} VAL very large data bases  COL year \textbf{time.event} VAL 2001 \\ \hline
    \end{tabular}
    }
\vspace{-0.3cm}
\end{table}

\textbf{Case Study III: Entity Resolution Results Predicted by Column-Level KA and Combined KA Methods are Different}

This case study compares the predicted results based on entry inputs augmented by column-level knowledge, i.p., by Doduo, and entry inputs augmented by combined knowledge, by Doduo and entity linking methods. These two records are incorrectly predicted as \textbf{not-match} by Doduo, but correctly predicted as \textbf{match} by combined knowledge. Initially, before knowledge injection, the two records seemed different due to distinct values in the author fields, and Doduo alone couldn't reflect the similarity. By adding the knowledge from entity linking, the dataset was enriched with additional labels such as ``scientist" for ``hector garcia-molina" and ``professional" for ``richard snodgrass". This additional layer of knowledge was able to improve the performance of the entity resolution model. 


\begin{table}[!ht]
    \centering
    \vspace{-0.3cm}
    \caption{Augmented Entity pairs at row 1457: The CST augmented by Doduo is highlighted in \textbf{bold}, and the knowledge augmented by entity linking (EL) is in \textcolor{red}{red}.}
    \vspace{-0.3cm}
    \label{tab:ea-3}
    \resizebox{\textwidth}{!}{
    \begin{tabular}{|p{2cm}|p{8cm}|p{8cm}|} \hline 
         Entity Pair &  entry\_doduo& entry\_doduo\_EL\\ \hline 
         Entity 1&  COL title \textbf{business.industry} VAL reminiscences on influential papers COL authors \textbf{people.person} VAL hector garcia-molina , patricia g. selinger , tomasz imielinski , david maier , jeffrey d. ullman , richard t. snodgrass COL venue \textbf{organization.organization} VAL sigmod record COL year \textbf{time.event} VAL 1998 & COL title \textbf{business.industry} VAL reminiscences on influential papers  COL authors \textbf{people.person} VAL hector garcia-molina \textcolor{red}{(scientist)} , patricia g. selinger , tomasz imielinski , david maier , jeffrey d. ullman , richard t. snodgrass  COL venue \textbf{organization.organization} VAL sigmod \textcolor{red}{(album)} record  COL year \textbf{time.event} VAL 1998 \textcolor{red}{(periodic process)} \\ \hline 
 
         Entity 2& COL title \textbf{business.industry} VAL reminiscences on influential papers COL authors \textbf{people.person} VAL richard snodgrass COL venue \textbf{book.periodical} VAL acm sigmod record COL year \textbf{time.event} VAL 1998 & COL title \textbf{business.industry} VAL reminiscences on influential papers  COL authors \textbf{people.person} VAL richard snodgrass \textcolor{red}{(professional)}  COL venue \textbf{book.periodical} VAL acm sigmod \textcolor{red}{(artificial physical object)} record  COL year \textbf{time.event} VAL 1998 \textcolor{red}{(periodic process)}\\ \hline
    \end{tabular}
    }
\vspace{-0.5cm}
\end{table}

\section{Conclusions}

We present T-KAER, a framework that enhances transparent entity resolution tasks by documenting the experimental process in log files. We conduct a case study on a citation dataset, both quantitative and qualitative analyses are processed on the log files. T-KAER allows us to address three transparency-related questions, elucidating: (1) How diverse entry inputs lead to variations in the performance of predicted results; (2) What distinct semantic information is contained in entry inputs; (3) How the internal products, specifically embeddings generated by pre-trained language models (PLMs), can accurately represent the semantic information derived from entry inputs.

\bibliographystyle{apalike}
\bibliography{ref}

\end{document}